\documentclass[letterpaper, 10 pt, conference]{ieeeconf} 
\IEEEoverridecommandlockouts                             
\usepackage{color}
\usepackage{graphicx} % for pdf, bitmapped graphics files
\usepackage{mathtools} 
\usepackage{siunitx}

\title{\LARGE \bf Self-Propelled Soft Everting Toroidal Robot\\ for Navigation and Climbing in Confined Spaces}

\author{Nelson G. Badillo Perez$^{1}$ and Margaret M. Coad$^{1}$
\thanks{$^{1}$Department of Aerospace and Mechanical Engineering, University of Notre Dame, Notre Dame, IN 46556, USA. {\tt\small \{nbadillo, mcoad\}@nd.edu}}%
}

\begin{document}

\maketitle
\thispagestyle{empty}
\pagestyle{empty}

\begin{abstract}

There are many spaces inaccessible to humans where robots could help deliver sensors and equipment. Many of these spaces contain three-dimensional passageways and uneven terrain that pose challenges for robot design and control. Everting toroidal robots, which move via simultaneous eversion and inversion of their body material, are promising for navigation in these types of spaces. We present a novel soft everting toroidal robot that propels itself using a motorized device inside an air-filled membrane. Our robot requires only a single control signal to move, can conform to its environment, and can climb vertically with a motor torque that is independent of the force used to brace the robot against its environment. We derive and validate models of the forces involved in its motion, and we demonstrate the robot's ability to navigate a maze and climb a pipe.

\end{abstract}

\section{Introduction}

Navigating confined spaces that are inaccessible to humans has long been a goal of robotics and a challenge for robot design and control. Inspection inside pipes~\cite{deepak2016development}, exploration in rubble~\cite{Whitman2018}, movement inside the human body~\cite{dupont2022continuum}, and monitoring in outdoor animal burrows~\cite{vercauteren2002camera} are examples of real-world scenarios where robots capable of confined-space navigation could be helpful. Often, these spaces contain three-dimensional passageways that are difficult to navigate for robots that are unable to brace themselves against the environment. In addition, these spaces may contain uneven terrain or blockages that require a robot to be able to change shape to conform to the environment.

One particularly promising method of confined-space robotic navigation is that of simultaneous eversion (i.e., turning inside-out) and inversion (i.e., turning outside-in) of the two ends of a toroidal membrane that makes up the robot body. This mechanism allows the outside of the robot body to remain stationary with respect to the environment while the inside cycles through and emerges from its tip, similar to a three-dimensional tank tread. Here, we refer to these robots as \textit{everting toroidal robots}, but they have also been called ``whole skin locomotion robots," ``hydrostatic skeleton crawlers," ``toroidal skin drive robots," and ``sliding membrane locomotion robots."

\begin{figure}[tb]
    \centering
    \includegraphics[width=\columnwidth]{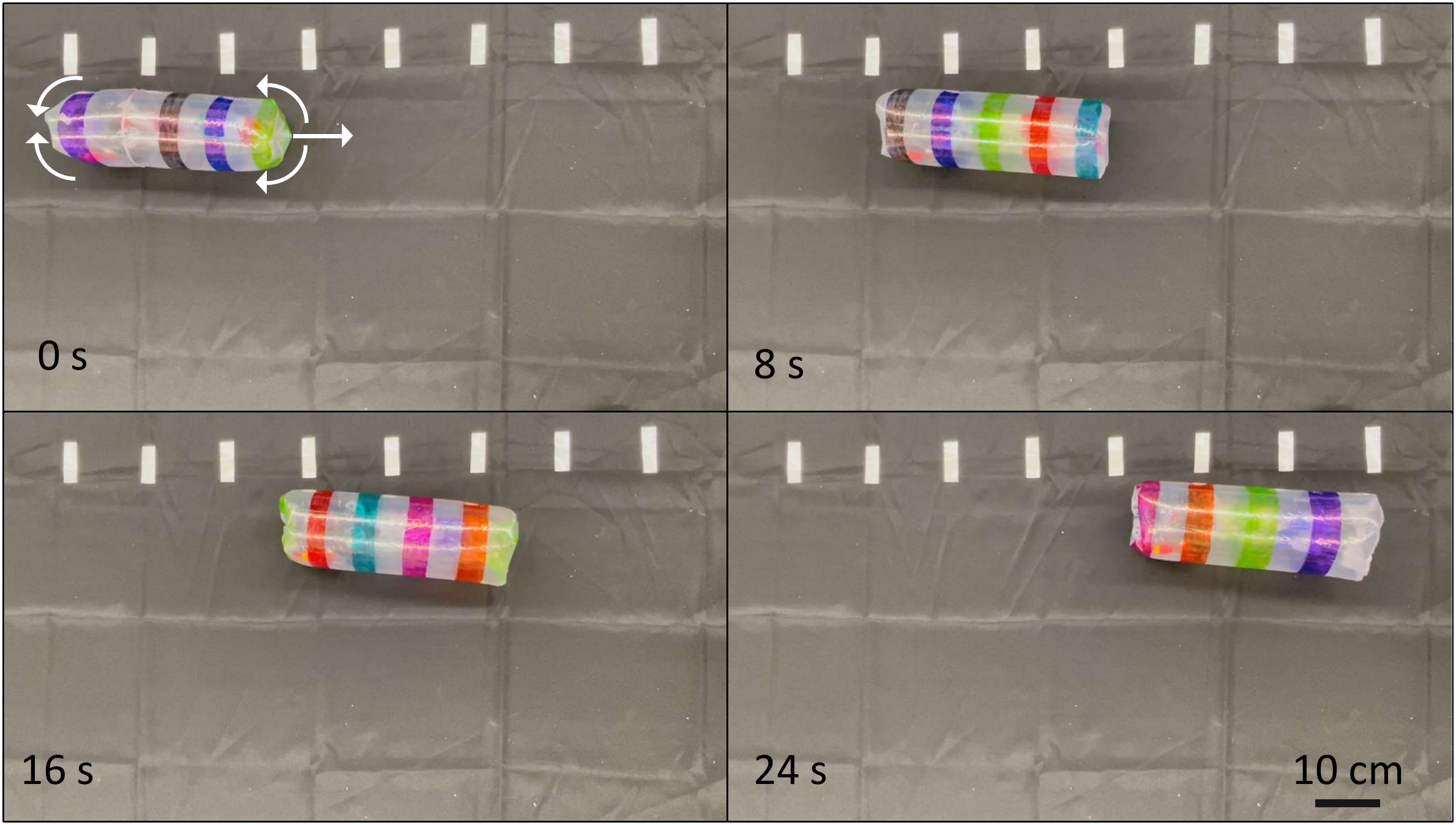}
    \caption{Our soft everting toroidal robot. Photos show the view from above at discrete time points during self-propelled locomotion of the robot along the floor. The robot moves by simultaneously everting its air filled membrane at the front end (on the right side in this image) and inverting the membrane at the back end (on the left side in this image) using an internal motorized device powered by a battery.}
    \label{fig:Overview}
    \vspace{-0.6cm}
\end{figure}

Early literature on everting toroidal robots developed concepts and prototypes of actuation techniques for propulsion of liquid-filled membranes, including contracting rings~\cite{ingram2006mechanics} and chemical membrane expansion~\cite{hong2009whole, orekhov2010mechanics, orekhov2010actuation}, but did not demonstrate a self-propelled robot. Kimura et al.~\cite{kimura2006flexible} demonstrated a self-propelled, air-filled everting toroidal robot, but it required complex coordination of the inflation and deflation of air pockets, and it did not fully recycle its membrane. More recently, Leon-Rodriguez et al.~\cite{leon2015ferrofluid} demonstrated locomotion of a ferrofluid-filled everting toroidal robot toward an external magnetic field, but this design could not propel itself in a non-magnetic environment. Various researchers~\cite{mckenna2008toroidal, molfino2008rescue, rimassa2009modular} have studied robots with a rigid internal skeleton that uses motorized rollers or worm gears to propel a toroidal skin, including a recent robot~\cite{consumi2022novel} that recycles tracks instead of a membrane and can adjust its diameter. A main challenge with these designs lies in the tradeoff between friction used to brace against the environment and friction that must be overcome to cycle the membrane. Finally, several recent designs~\cite{li2020bioinspired, sui2020bioinspired, root2021bio} have been developed for liquid-filled soft everting toroidal grippers that are driven by mechanical plungers, and a liquid-filled toroidal robot driven by a string~\cite{takahashi2020retraction}, but these robots cannot propel themselves or continuously recycle their membrane.

Another related class of robots is everting vine robots~\cite{blumenschein2020review}, which use air-pressure-driven eversion to lengthen from the tip and navigate their environment. These robots have shown benefits for navigating confined spaces, such as the ability to squeeze their body through small apertures~\cite{HawkesScienceRobotics2017}, bend to navigate around obstacles in their environment~\cite{GreerICRA2018}, and support their body weight to travel up and over obstacles~\cite{HawkesScienceRobotics2017}. Compared to everting vine robots, everting toroidal robots have the additional benefit that they can move unconstrained by a connection to a base, which eliminates the path-dependence of vine robot propulsion forces~\cite{blumenschein2017modeling} and the need to fabricate long lengths of material for the robot body.

\begin{figure*}[tb]
    \centering
    \includegraphics[width=\textwidth]{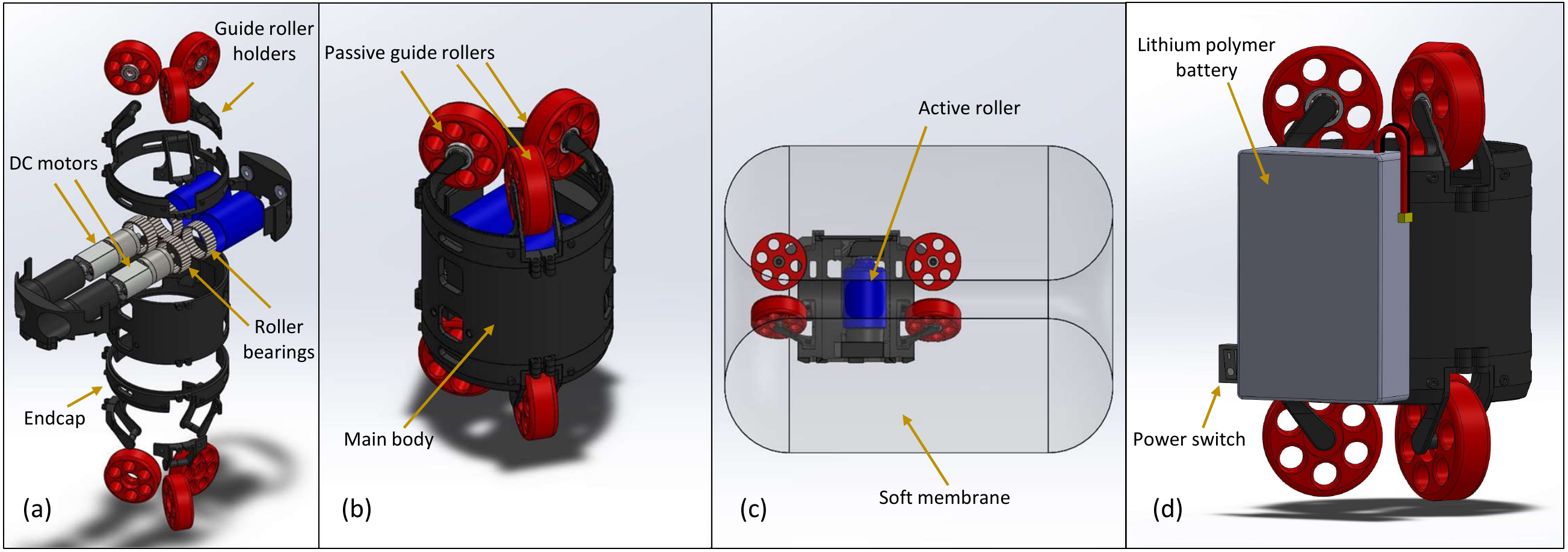}
    \caption{Design of our propulsion device. (a) Exploded view showing the internal DC motors and roller bearings which drive the device. (b) Isometric view showing the passive rollers for grounding the device against the membrane end and guiding the membrane tail into the device. (c) Side view showing the device inside the soft membrane, with the device active rollers gripping the membrane tail. (d) Side view showing the lithium polymer battery and power switch attached to the main body of the propulsion device.}
    \label{fig:Design}
    \vspace{-0.5cm}
\end{figure*}

In this paper, we present a novel self-propelled soft everting toroidal robot (Fig.~\ref{fig:Overview}) that continuously recycles an air-filled membrane using a motorized device that sits inside the pressurized part of the robot body. Our robot requires only a single control signal to move, can conform to obstacles in its environment, and can climb vertically with a motor torque that is independent of the force used to brace the robot against its environment. Our work uses design insights from everting vine robots, namely the retraction device presented in~\cite{CoadRetraction2020,JeongIROS2020}, to design the propulsion mechanism. The remainder of this paper discusses the design and fabrication of our robot (Section~\ref{sec:Design}), modeling and experimental validation of the forces involved in its movement (Section~\ref{sec:Modeling}), and demonstration of its ability to navigate and climb in confined spaces (Section~\ref{sec:Demonstration}).

\section{Design and Fabrication} \label{sec:Design}

Soft everting toroidal robots present a unique set of design considerations. We present these considerations in this section, along with the design and fabrication of our robot.

\subsection{Design Considerations}

Our goal was to create a robot that can locomote without sliding of its body relative to its environment. To further enhance the robot's adaptability to unpredictable environments, we chose to make the robot out of a compliant membrane and to minimize the size of the rigid components. To provide structure to the membrane, we chose to fill it with air as the working fluid, because air is readily available and can be easily input into the membrane with a traditional air pump. Liquids could also be used as the working fluid, but any internal electronics would need to be protected.

To evert and invert the air-filled membrane and achieve locomotion, we needed to develop a propulsion mechanism for the robot. We chose to base our design on the vine robot retraction mechanism presented in~\cite{CoadRetraction2020, JeongIROS2020}, which has a very similar purpose. This actuation mechanism operates by creating a force between one end of the membrane and the inner part of the membrane (the ``tail"), which causes inversion at that end of the robot. Like the devices in~\cite{CoadRetraction2020, JeongIROS2020}, our design creates this force based on torque from a pair of motor-driven rollers that grip the tail. This choice of actuation scheme only requires a single control input to continuously recycle the membrane. Reversing the polarity of the motor voltage in the propulsion device allows the robot to drive in the opposite direction, but here we chose to run our prototype in only one direction to demonstrate the most basic functionality.

\subsection{Propulsion Device}
Fig.~\ref{fig:Design} shows the design of our propulsion device. The structural components (black) consist of a main body that holds the motor housing assembly, twelve passive roller holders which restrict the translational motion while enabling the rotational motion of the passive rollers, and two endcaps that fasten the passive roller holders to the main body (Fig.~\ref{fig:Design}(a)). At each end of the device are three passive rollers (red) that are used to exert a grounding force on the end of the membrane and are fitted with two ball bearings each to minimize the potential for twisting (Fig.~\ref{fig:Design}(b)). The middle of the device contains two active rollers (blue) that are each driven by a DC motor (3485, Pololu Corporation, Las Vegas, NV). The rollers connect to the motors using set screws and roll on the housing using roller bearings. The rollers grip the membrane tail to move the robot (Fig.~\ref{fig:Design}(c)). We 3D printed the components of the propulsion device using polylactic acid (PLA), and we assembled the components using screws and nylon locking nuts, which allows for the quick servicing of parts. For the fully portable version of the device, a lithium polymer battery (RDQ5000-7.4-5, Race Day Quads, Orlando, FL) powers the device, and the voltage to the motors can be turned on with a power switch (Fig.~\ref{fig:Design}(d)). The weight of the propulsion device without the battery is 360~g, and the weight with the battery is 574~g.
% Change made above%

\subsection{Membrane}

\begin{figure}[tb]
    \centering
    \includegraphics[width=\columnwidth]{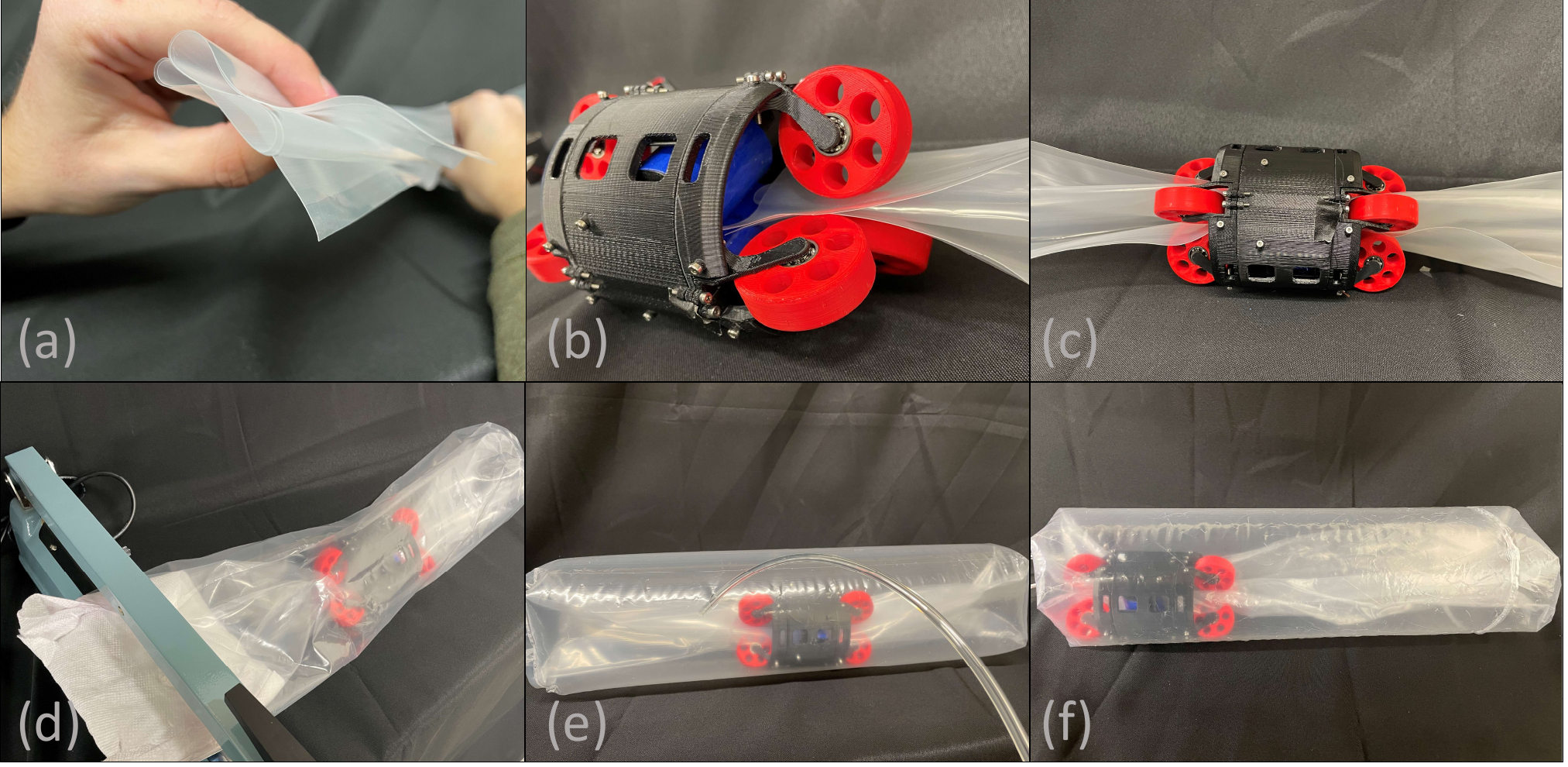}
    \caption{Fabrication of soft membrane and placement of the propulsion device inside. (a) Step 1: Fold the membrane into 4 layers. (b) Step 2: Insert the tip of the membrane into the propulsion device rollers. (c) Step 3: Turn on the device and pull through two-thirds of the membrane. (d) Step 4: Heat seal the membrane with a paper towel inside to achieve a toroidal shape. (e) Step 5: Fill the membrane with air. (f) Step 6: Tape the air hole to create an air-tight seal.}
    \vspace{-0.5cm}
    \label{fig:Fabrication}
\end{figure}

The membrane of the robot is made of low-density polyethylene (LDPE) plastic. We chose this material for its high compliance, low cost, airtightness, and ability to be sealed using an impulse heat sealer. Fig.~\ref{fig:Fabrication} shows the process used to fabricate the membrane into a toroidal shape and place the propulsion device inside it. Unless otherwise noted, we used membranes 4~mm thick and 13.7~cm in inflated outer diameter for all experiments. First, we folded the membrane into four layers (Fig.~\ref{fig:Fabrication}(a)), and then we placed the membrane tip inside the propulsion device (Fig.~\ref{fig:Fabrication}(b)). Then, we turned on the device and sucked the membrane into it by two-thirds of its length (Fig.~\ref{fig:Fabrication}(c)). Then, we folded the longer end of the membrane inside-out and pulled it over the device until it reached the other end. From there, we placed a paper towel inside the end of the innermost membrane and heat sealed that end (Fig.~\ref{fig:Fabrication}(d)). The paper towel prevents the inner membrane from being sealed to itself while allowing the inner and outer membrane to be sealed together, thus achieving a toroidal shape. After removing the paper towel and cutting off the excess membrane material, we cut a small hole in the membrane and filled it with air (Fig.~\ref{fig:Fabrication}(e)). Lastly, we removed the air tube and sealed the hole with a small piece of tape (Fig.~\ref{fig:Fabrication}(f)). This allows for the membrane to maintain pressure when desired and be easily emptied later.

\section{Modeling and Experimental Validation} \label{sec:Modeling}

To understand the forces involved in the robot's movement, we derived analytical models based on force balance of the membrane and propulsion device. Here, we present these models and their experimental validation.

\subsection{Quasistatic Force Modeling}

\begin{figure}[tb]
    \centering
    \includegraphics[width=\columnwidth]{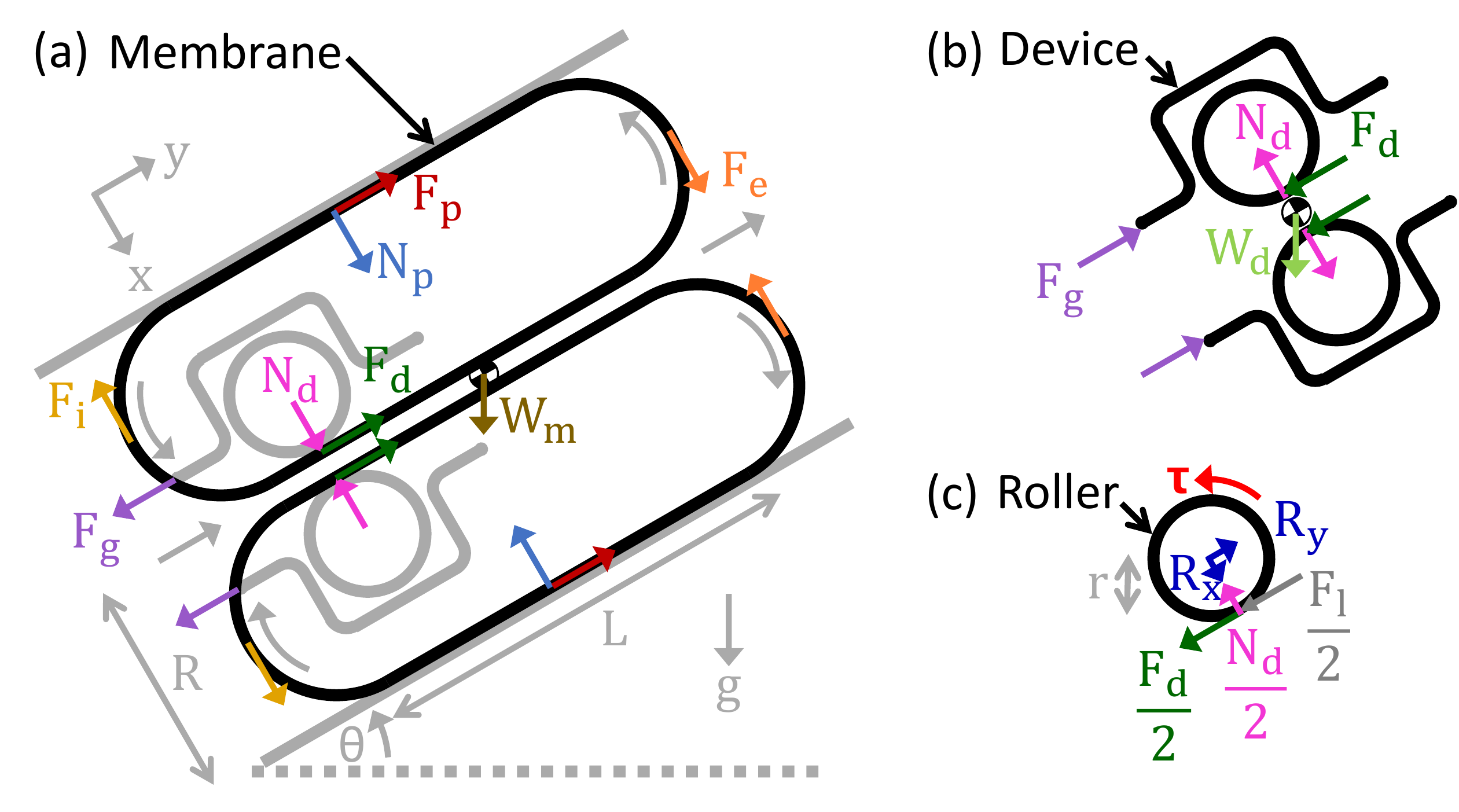}
    \caption{Free body diagrams of the robot while moving from lower left to upper right inside a pipe at an angle $\theta$ above the horizontal. (a) Forces acting on the membrane. (b) Forces acting on the propulsion device. (c) Forces acting on one of the two actively driven device rollers.}
    \vspace{-0.5cm}
    \label{fig:FBD}
\end{figure}

Fig.~\ref{fig:FBD} shows free body diagrams of the robot climbing up a pipe oriented at an angle $\theta \in [-90^\circ, 90^\circ]$ above horizontal. The $x$ axis is perpendicular to the direction of motion, and the $y$ axis is in the direction of motion. Fig.~\ref{fig:FBD}(a) shows the forces acting on the membrane, which include the weight $W_m$ of the membrane, as well as forces due to the membrane's contact with the pipe (the normal force $N_p$ acting inward on the membrane wall from the pipe, and the frictional force $F_p$ acting on the membrane wall from the pipe around its circumference). Three additional forces act on the membrane due to the contact between the membrane and the propulsion device (the normal force $N_d$ acting on the membrane tail from the propulsion device, the frictional force $F_d$ acting on the membrane tail from the propulsion device, and the grounding force $F_g$ applied on the inverting end of the robot by the propulsion device). We also include a force $F_i$ at the inverting end of the robot that opposes the direction of inversion, and a force $F_e$ at the everting end of the robot that opposes the direction of eversion; these forces were shown to exist for eversion in~\cite{HawkesScienceRobotics2017} and for inversion in~\cite{CoadRetraction2020} and represent the force required to wrinkle and unwrinkle the membrane at its tip. Note that air pressure due to the pressure differential between the inside and outside of the membrane acts in all directions perpendicular to the membrane, but it does not exert a net force on the membrane, because the membrane is a closed volume. 

Fig.~\ref{fig:FBD}(b) shows the forces acting on the propulsion device, which include the weight $W_d$ of the propulsion device, as well as the forces due to contact with the membrane (normal force $N_d$, frictional force $F_d$, and grounding force $F_g$). Finally, Fig.~\ref{fig:FBD}(c) shows the forces acting on a single roller of the propulsion device, which include half of the device normal force $N_d$, half of the device frictional force $F_d$, reaction forces $R_x$ and $R_y$ at the joint that holds the roller in place, torque $\tau$ applied by the motor on the roller, and half of the quantity $F_l$, which represents the losses in force between the output shafts of the motors and the membrane due to factors such as misalignment between the motor shafts and the rollers, and friction in the roller bearings.

To derive equations relating the forces acting on the robot, we begin by setting the forces acting on the membrane in the $y$ direction to zero, since the robot moves quasistatically during climbing:
\begin{eqnarray}\label{eqn:membrane_y_force}
F_p + F_d - W_m \sin{\theta} - F_g = 0.
\end{eqnarray}
Next, we conduct a tension balance along the membrane, which considers the forces acting along the entire membrane and sums them to zero:
\begin{eqnarray}\label{eqn:membrane_tension}
F_e + F_i + F_p - F_d = 0.
\end{eqnarray}
Finally, we set the forces acting on the propulsion device in the $y$ direction to zero:
\begin{eqnarray}\label{eqn:device_y_force}
F_g - F_d - W_d \sin{\theta} = 0.
\end{eqnarray}

\begin{figure*}[ht]
    \centering
    \includegraphics[width=0.8\textwidth]{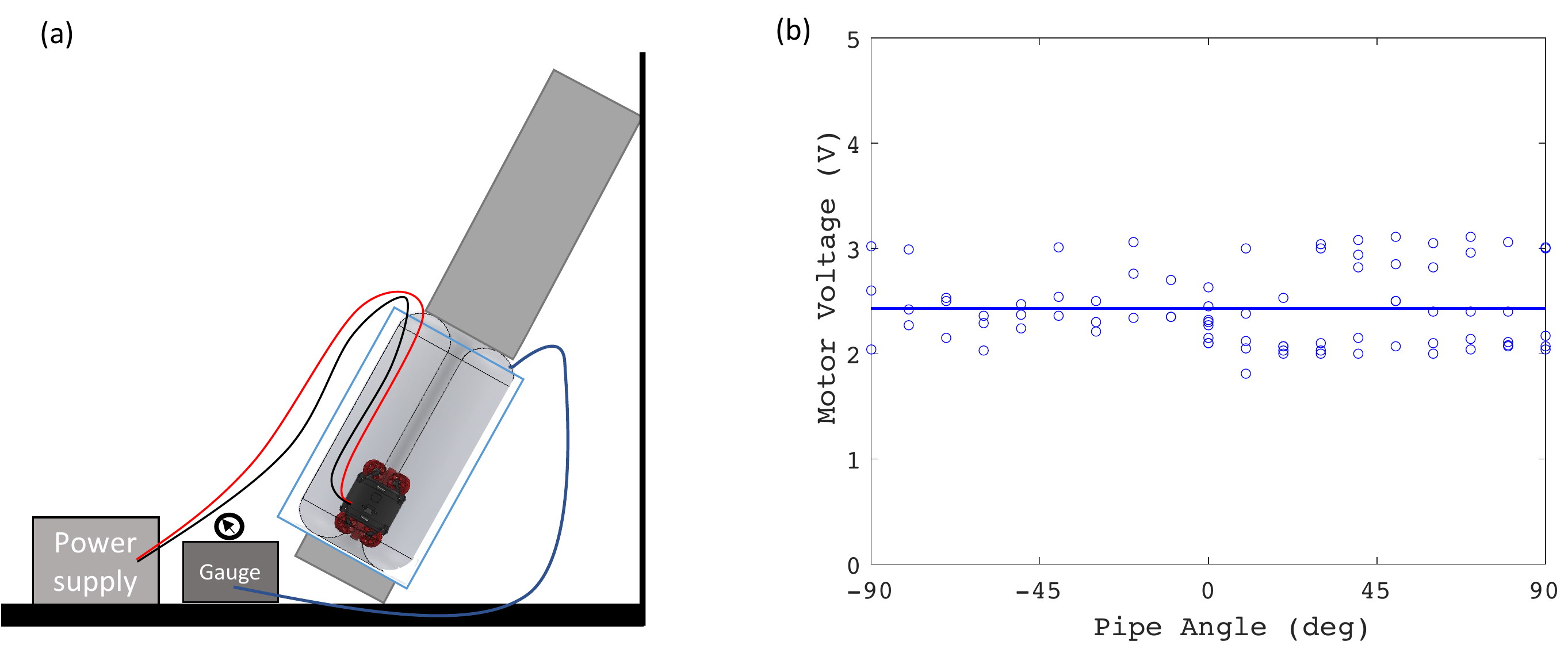}
    \caption{Propulsion force validation experiment. (a) Experimental setup used to measure the minimum propulsion device motor voltage required to begin climbing a pipe at different angles. The setup includes a wood plank, acrylic pipe, power supply, pressure gauge, and the robot. (b) Experimentally determined values of the minimum motor voltage required to climb up or down the pipe. Positive pipe angles denote that the robot is climbing upwards, and negative pipe angles denote that the robot is climbing downwards. The circles are measured data points, and the solid curve is the modeled voltage (Eqn.~\ref{eqn:angle_voltage_relationship}), which varies from 2.41~V at -90 degrees to 2.45~V at 90 degrees.}
    \vspace{-0.7cm}
    \label{fig:PropulsionForce}
\end{figure*}

Using Eqns.~\ref{eqn:membrane_y_force}, \ref{eqn:membrane_tension}, and~\ref{eqn:device_y_force}, we can solve for the unknowns $F_d$, $F_g$, and $F_p$ in terms of the constants $F_e$, $F_i$, $W_m$, and $W_d$. Solving for $F_d$, we have
\begin{eqnarray}\label{eqn:F_d}
F_d = F_e+F_i+(W_m+W_d)\sin{\theta}.
\end{eqnarray}
Plugging this expression for $F_d$ into Eqn.~\ref{eqn:device_y_force}, we have
\begin{eqnarray}\label{eqn:F_g}
F_g = F_e+F_i+(W_m+2W_d)\sin{\theta}.
\end{eqnarray}
Plugging the expression for $F_d$ into Eqn.~\ref{eqn:membrane_tension}, we have
\begin{eqnarray}\label{eqn:F_p}
F_p = (W_m+W_d)\sin{\theta}.
\end{eqnarray}

Eqns.~\ref{eqn:F_d},~\ref{eqn:F_g}, and~\ref{eqn:F_p} hold for positive, zero, and negative values of $\theta$ (i.e., climbing up, horizontally, or down). Note that, for negative $\theta$, Eqn.~\ref{eqn:F_p} yields a negative value for $F_p$, denoting that it acts in the opposite direction to how it is drawn in Fig.~\ref{fig:FBD}(a). Also note that, for a negative $\theta$ value of magnitude large enough that the magnitude of $(W_m+2W_d) \sin{\theta}$ is greater than the sum of $F_e$ and $F_i$, Eqn.~\ref{eqn:F_g} yields a negative value for $F_g$. In this case, the propulsion device will drive itself within the membrane until it is contacting the everting (i.e., the downward) end of the membrane. There, it will apply its grounding force in the opposite direction as $F_g$ is drawn in Fig.~\ref{fig:FBD}(a).

It can be useful for understanding motor requirements, and for device force validation, such as in Sec.~\ref{sec:Modeling}B, to relate the voltage $V$ applied to the propulsion device motors to the device force $F_d$ and climbing angle $\theta$. From Fig.~\ref{fig:FBD}(c), we can begin by relating the motor torque $\tau$ applied on a single roller of radius $r$ to the device force $F_d$. Setting the sum of the moments applied on the roller about its center in the $z$ direction equal to zero and solving for $\tau$, we have:
\begin{eqnarray}\label{eqn:roller_moment}
\tau = \dfrac{r}{2}(F_d+F_l).
\end{eqnarray}
Motor torque $\tau$ for each DC motor can be modeled as a function of current $I$ and torque constant $K_\tau$:
\begin{eqnarray}\label{eqn:torque_constant}
\tau = K_{\tau}I.
\end{eqnarray}
Current $I$ through each motor while stalled at steady state can be represented in terms of the voltage $V$ across its leads and the motor resistance $R$:
\begin{eqnarray}\label{eqn:current_voltage_resistance}
I = \dfrac{V}{R}.
\end{eqnarray}
Plugging Eqn.~\ref{eqn:current_voltage_resistance} into Eqn.~\ref{eqn:torque_constant}, and plugging Eqn.~\ref{eqn:torque_constant} into Eqn.~\ref{eqn:roller_moment}, and solving for the motor voltage $V$, we have the following relation between voltage and device force at stall:
\begin{eqnarray}\label{eqn:device_voltage_relationship}
V = \dfrac{rR}{2K_{\tau}}(F_d+F_l).
\end{eqnarray}
Finally, plugging Eqn.~\ref{eqn:F_d} into Eqn.~\ref{eqn:device_voltage_relationship} and grouping terms based on dependence on the pipe angle $\theta$, we have the following relation between voltage and pipe angle:
\begin{eqnarray}\label{eqn:angle_voltage_relationship}
V = \dfrac{rR}{2K_{\tau}}(F_e+F_i+F_l)+\dfrac{rR}{2K_{\tau}}(W_m+W_d)\sin{\theta}.
\end{eqnarray}

Now we consider the friction force $F_p$ between the membrane wall and the pipe that is required for the robot to hold itself inside a pipe without slipping. Based on the law of static friction, we expect that the robot will start to fall when the the coefficient of static friction $\mu_s$ times the normal force $N_p$ between the membrane wall and the pipe equals the force $F_p$ to support the robot's weight; larger values of $N_p$ will ensure that the robot does not slip. Assuming that the outer diameter of the membrane is larger than the inner diameter of the pipe, part of this normal force is generated by the membrane's internal pressure $P$ multiplied by the contact area between the membrane and the pipe, which equals the length $L$ of the part of the robot in contact with the pipe multiplied by the pipe inner circumference $2 \pi R$, where $R$ is the pipe inner radius. The rest of the normal force is generated by opposing the portion of the total robot weight that is perpendicular to the pipe wall:
\begin{eqnarray}\label{eqn:slip_force}
F_p = \mu_s N_p = \mu_s [P (2 \pi R L)+(W_m+W_d) \cos{\theta}].
\end{eqnarray}
In order to determine the maximum robot weight that can be supported in a vertical pipe (such as in Sec.~\ref{sec:Modeling}C), we can set $F_p$ in Eqns.~\ref{eqn:F_p} and~\ref{eqn:slip_force} equal, with $\theta$ equal to 90$^\circ$, and solve for the weight:
\begin{eqnarray}\label{eqn:slip_weight}
W_m + W_d = \mu_s [P (2 \pi R L)].
\end{eqnarray}

\subsection{Propulsion Force Validation}

To experimentally validate our model for $F_d$ (Eqn.~\ref{eqn:F_d}), we conducted an experiment (Fig.~\ref{fig:PropulsionForce}) where the robot climbed up or down a cylindrical pipe at varying angles, and we measured the motor voltage required to begin moving. This motor voltage could be considered a proxy for $F_d$, because the two are linearly proportional based on motor and roller constants with an offset for force losses (Eqn.~\ref{eqn:device_voltage_relationship}). The resulting expected relationship between motor voltage and pipe angle is given in Eqn.~\ref{eqn:angle_voltage_relationship}, where $r$, $R$, $K_\tau$, $W_m$, and $W_d$ are measurable constants, and $F_e$, $F_i$, and $F_l$ are unknowns.

The experimental setup (Fig.~\ref{fig:PropulsionForce}(a)) consisted of an acrylic pipe of inner diameter 12.4~cm and length 30.5~cm fastened to a wooden plank which could be placed at a desired angle. In order to easily vary the motor voltage, we powered the propulsion device via wires that passed through a hole in the membrane and connected to an external power supply. We placed the robot inside the pipe and used a manually controlled pressure regulator and gauge to maintain a constant pressure. We moved the pipe through angles at 10$^\circ$ intervals ranging from 0$^\circ$ to 90$^\circ$ above the horizontal while the robot climbed up the pipe, and then again while the robot climbed down. For climbing up, we repeated the experiment for five trials, and for climbing down, we repeated the experiment for three trials (fewer than for climbing up, because we expected to see symmetrical results). At each angle, we increased the power supply voltage until the robot began moving and recorded that voltage.

Fig.~\ref{fig:PropulsionForce}(b) shows the results. The trials where the robot is climbing up are plotted as 0$^\circ$ to 90$^\circ$ and climbing down from 0$^\circ$ to -90$^\circ$. The results show that the motor voltage varied between trials by approximately 1~V, likely due to variations in the inversion/eversion of the membrane and its feeding into the rollers. The results also showed that there was no clear dependence of the motor voltage on the pipe angle. The average voltage across all trials was 2.43~V. We also repeated these experiments at a range of pressures and did not observe a change in the results.

In order to plot Eqn.~\ref{eqn:angle_voltage_relationship} alongside the data, we took the average voltage across all trials to be equal to the first term in Eqn.~\ref{eqn:angle_voltage_relationship}, i.e., the voltage when $\theta$ equals zero. To calculate the second term, we calculated/measured the relevant constants. We measured the roller radius $r$ to be 0.017~m. We used Eqn.~\ref{eqn:current_voltage_resistance} to calculate that the motor resistance $R$ is 7.5~$\Omega$ by plugging in the rated motor voltage of 12~V and the rated current at stall of 1.6~A. We used Eqn.~\ref{eqn:torque_constant} to calculate that $K_\tau$ equals 1.53~$\frac{Nm}{A}$ by plugging in the motor's listed stall torque (25~kg-cm) and stall current (1.6~A) at 12~V. We measured the weight of the membrane $W_m$ to be 85~g, and the weight of the propulsion device $W_d$ to be 360~g. Using these values, we plotted our model for voltage as a function of pipe angle (Eqn.~\ref{eqn:angle_voltage_relationship}) as the blue curve in Fig.~\ref{fig:PropulsionForce}(b). 

As shown in both the model and the data, for the robot weight tested here, the magnitude of the eversion and inversion forces $F_e$ and $F_i$ combined with the force losses $F_l$ is much larger than the magnitude of the weight $W_m+W_d$, such that the difference between adding and subtracting the robot weight to and from $F_e + F_i + F_l$ is masked by the variability in motor voltage between trials. Quantifying the relative values of $F_e$, $F_i$, and $F_l$ is left to future work, but using Eqn.~\ref{eqn:angle_voltage_relationship} with $\theta$ equal to zero and $V$ equal to 2.43~V, we can solve for their sum to find that $F_e + F_i + F_l$ is equal to 58.3~N, and we know that $W_m + W_d$ equals 4.4~N. Based on~\cite{CoadRetraction2020}, which measured the inversion force $F_i$ for an LDPE membrane of similar diameter and thickness to be 7.0~N, we can approximate that $F_e$ is likely around 10~N, $F_i$ is likely also around 10~N, and $F_l$ is likely around 40~N.

These results are interesting, as they indicate that force losses between the motor output shaft and the membrane make up a significant portion of the voltage required by the device motors, which means that the dependence of the operating voltage on the robot weight and thus the angle of propulsion is very small. If the robot weight increases significantly (e.g., to 10x or 100x the current weight), the dependence of the device propulsion force on the pipe angle will become significant, but for our current robot design, the robot can easily propel itself at any angle without an appreciable change in operating voltage.

\subsection{Slipping Force Validation}

\begin{figure}[tb]
    \centering
    \includegraphics[width=\columnwidth]{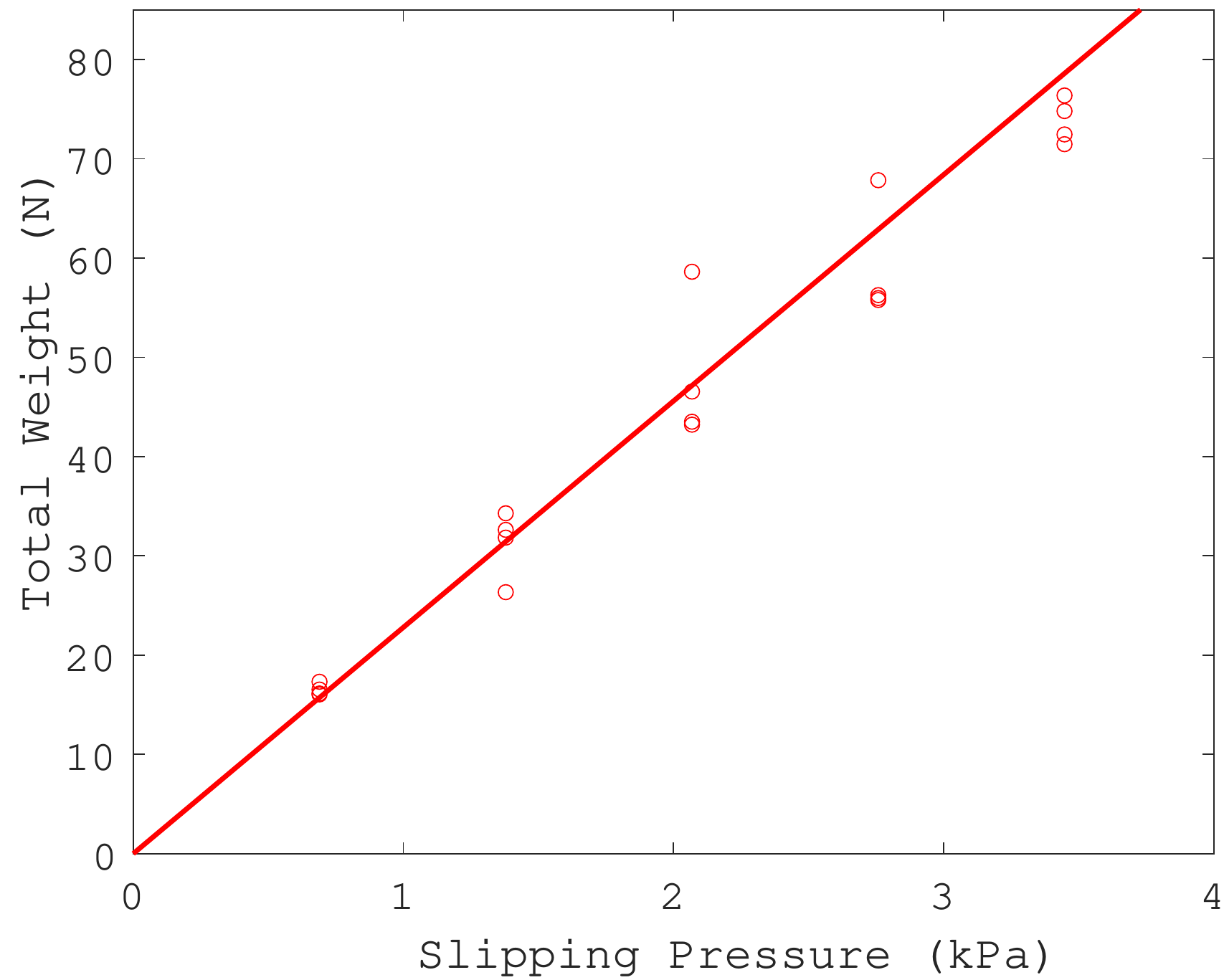}
    \caption{Slipping force validation experiment. Data shows the minimum value of the total weight of the membrane plus an externally applied load required for the robot to begin slipping at various values of internal robot pressure. The circles are measured data points, and the solid line is the model from Eqn.~\ref{eqn:slip_weight}.}
    \vspace{-0.6cm}
    \label{fig:SlippingForce}
\end{figure}

To experimentally validate the relationship between slipping force, air pressure, and pipe geometry, we conducted a set of experiments where an inflated membrane was placed inside a vertical acrylic pipe, and we measured the minimum applied load required to cause it to begin slipping at various pressures.

We first determined the coefficient of static friction $\mu_s$ by placing an LDPE membrane on top of an acrylic panel with a weight on top. We used a digital scale (ES-PS01, Dr. Meter) to measure the force needed to begin moving the membrane and compared it with the known value of the weight. We collected and averaged five values to find that $\mu_s$ is approximately 0.192.

For validating the model, we used a 19.4~cm inflated outer diameter LDPE membrane sealed to form a tube. The weight of the membrane was 85~g. We placed the membrane inside the same pipe as in the previous subsection held vertically and wrapped a high strength fishing string around the membrane vertically. The membrane was longer than the pipe, so we used the pipe length as the length of contact. We then inflated the membrane using a pressure regulator (QB3, Proportion-Air, McCordsville, IN) at five values ranging from 0.70 kPa to 3.45 kPa. We attached the digital scale to the bottom of the string and measured the force that caused the membrane to fall at each pressure. We completed four trials for each pressure. Fig.~\ref{fig:SlippingForce} shows the experimental results for the measured applied load plus the membrane weight, along with our model (Eqn.~\ref{eqn:slip_weight}), which matches well.

These results show the capability of our robot to support significant weight without slipping while climbing a pipe. Even for our relatively slippery membrane and pipe material, the minimum internal pressure required to support the robot weight with the battery is only 0.3~kPa. By increasing the internal pressure to 3.45~kPa, the robot can support approximately 80~N, and it can support heavier loads up to the burst pressure of the membrane, which should be on the order of 10~kPa. Higher loads can be supported by increasing the burst pressure of the membrane, the coefficient of friction between the membrane and the pipe, and/or the contact area between the membrane and the pipe.

\section{Demonstration} \label{sec:Demonstration}

We conducted a set of demonstrations to showcase the locomotive and compliant functionality of the everting toroidal robot. These demonstrations could be considered as mock scenarios similar to what might be seen in a search and rescue mission, such as navigating through collapsed structures horizontally and vertically. In these demonstrations, the robot successfully traverses a zigzagging maze with a small aperture, and it climbs up a pipe.

\subsection{Maze}

\begin{figure*}[bt]
    \centering
    \includegraphics[width=\textwidth]{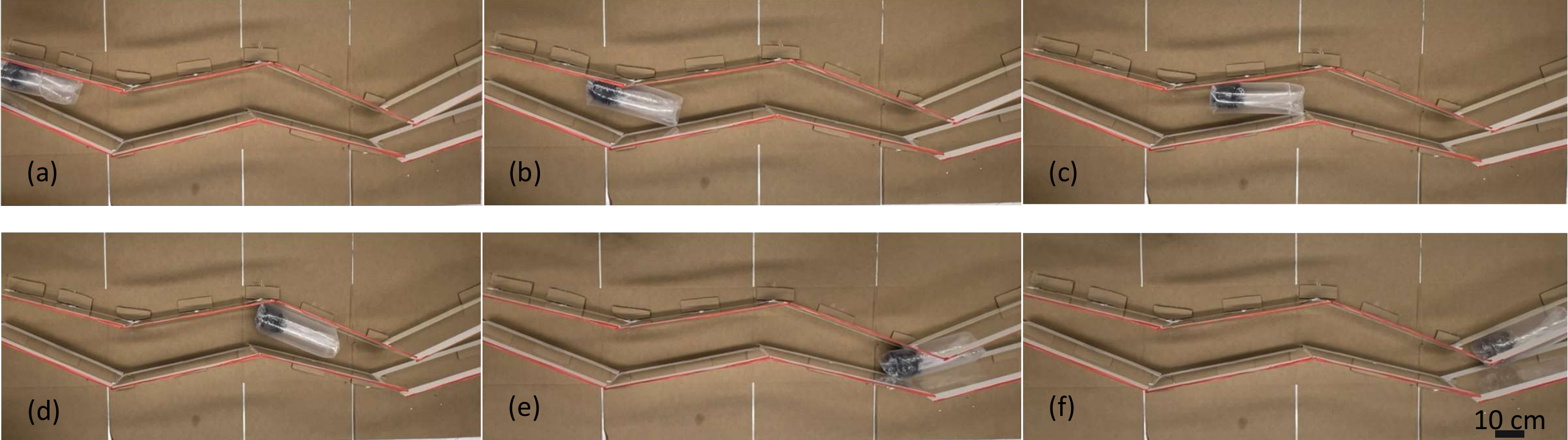}
    \caption{Demonstration of our soft everting toroidal robot navigating a maze. Photos show the view from above as the robot propels itself horizontally, (a) beginning at the left end of the maze, (b-c) moving along the lower maze wall, (d) moving along the upper maze wall, and finally (e) compressing through an aperture smaller than its membrane diameter to (f) reach the end of the maze.}
    \vspace{-0.3cm}
    \label{fig:Maze}
\end{figure*}

For the maze demonstration, we constructed a maze by fixing acrylic panels to a cardboard base. The panels form a zigzagging path with variable spacing between them. As shown in Fig.~\ref{fig:Maze}, the robot successfully traversed the maze. As the robot passes through the maze, it is forced to change direction multiple times as it comes into contact with a wall. Due to the robot's natural compliance and its unique eversion mechanism, the robot is able to turn to move along the wall, without any active steering (Fig.~\ref{fig:Maze}(a-d)). At the end of the maze (Fig.~\ref{fig:Maze}(e-f)), the robot encounters an aperture of width 11~cm, which is smaller than the diameter of the membrane (13.7~cm) but larger than the diameter of the propulsion device with the battery (10.4~cm). To navigate the aperture, the membrane squeezes laterally and elongates vertically.

\subsection{Pipe}

\begin{figure}[tb]
    \centering
    \includegraphics[width=0.6\columnwidth]{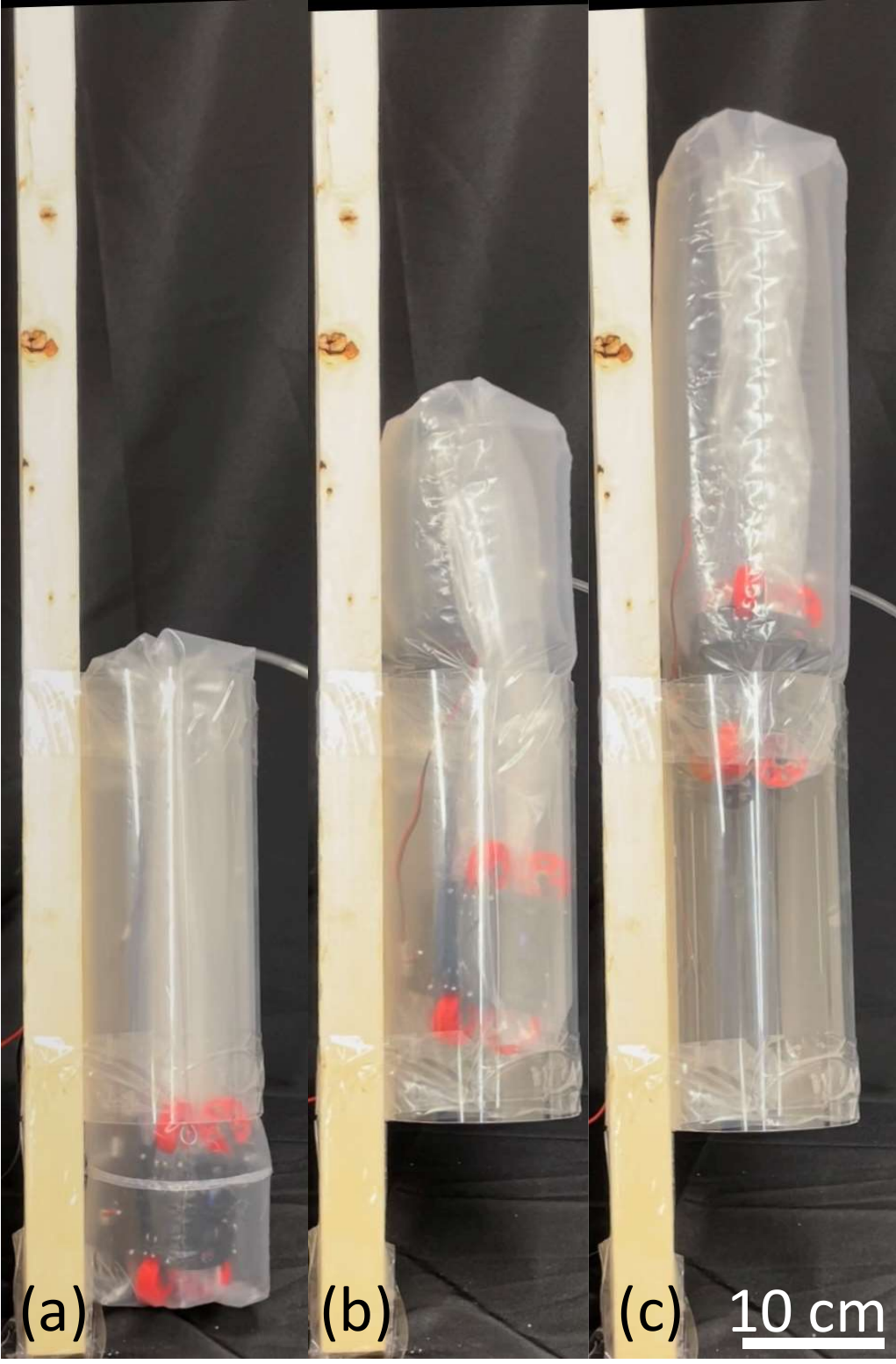}
    \caption{Demonstration of our soft everting toroidal robot climbing up a vertical pipe. Photos show the view from the side as the robot propels itself upwards, (a) beginning at the initial position at t = 0~s with the propulsion device below the pipe, (b) moving at t = 2.5~s half-way up the pipe, and (c) ending at t = 5~s at the top of the pipe.}
    \vspace{-0.5cm}
    \label{fig:Pipe}
\end{figure}

For the pipe climbing demonstration, we fastened an acrylic pipe to a wood plank and placed the robot inside the pipe. As shown in Fig.~\ref{fig:Pipe}, the robot successfully climbed the pipe. In its initial position, the propulsion device is placed below the pipe (Fig.~\ref{fig:Pipe}(a)). Then, the robot is powered on and climbs the pipe in approximately 5 seconds (Fig.~\ref{fig:Pipe}(b-c)).

% \section{Discussion}

% Despite the significant promise of this robot, there are several challenges that were encountered through its development. As a result of its unique toroidal geometry achieving an air tight sealed membrane is quite difficult and requires several attempts before each demonstration. It is quite common for the heat sealer used to not apply heat uniformly and holes would have to be taped to maintain constant pressure. The membrane would also wrinkle at the point of sealing and the extra material would not allow for an airtight seal. Membrane positioning was also a significant challenge particularly when the robot would have the battery attached. As a result of the weight imbalance of the robot with the battery, the propulsion device would twist the membrane while travelling and eventually cause the robot to stop. The current design also does not allow for the membrane to remain sealed if the propulsion device needs servicing or the battery needs recharging. 

% As the design of this robot progresses through future work, these issues will get resolved. Future challenges that we expect to arise are attaching sensors and cameras to the robot. Due to the everting mechanism of this robot it will be difficult to mount any device to the outside of the membrane.

% In our current design, the membrane needs to be completely airtight for the robot to operate without losing pressure, but in future iterations, version of the robot could be build that can actively let air in and out of the membrane.

\section{Conclusion and Future Work}

We presented a novel self-propelled soft everting toroidal robot that can navigate confined spaces. Our robot requires only a single control signal to move and can conform to its environment using its soft body. We presented the design considerations and manufacturing steps for the motorized propulsion device and inflatable membrane. We demonstrated that this robot can successfully navigate a cluttered environment, fit through an aperture, and climb a pipe. Our mathematical models and experimental validation showed that the propulsion force is independent of the force used to brace the robot against a pipe wall, and that, for relatively small robot weights, it has almost no dependence on the climbing angle. We also showed that the robot can support significant weight without slipping when climbing a pipe.

Future work will develop mechanisms for active steering and adjustment of the length and diameter of the robot, and methods to attach cameras and sensors. We will also study the robot's controllability and methods for wireless operation. We will explore the curvatures and paths this robot can successfully traverse, and the effect of different membrane materials, lengths, diameters, and internal pressures on locomotion. Additionally, we will explore the application of this robot for use in endoscopies and pipe inspection.

%Future work will develop mechanisms for active steering and active adjustment of the length and diameter of the robot. We will also explore different membrane materials and methods to add cameras and sensors to the robot.%

\section{Acknowledgements}

The authors thank the Dept. of Aerospace and Mechanical Engineering for their support, Fabely Moreno for assistance in the demonstrations, and Mark Yim for useful discussions.

% We would like to thank the Department of Aerospace and Mechanical Engineering of the University of Notre Dame for their continued support in this project. We would also like to acknowledge the contributions of University of Notre Dame student Fabely Moreno for her assistance in several demonstration tasks. We also thank Dr. Mark Yim at the University of Pennsylvania for his knowledge and input to the project.
% Talk about making the inner device smaller.

\bibliographystyle{IEEEtran}
\bibliography{library}

\end{document}